%% file: main.tex
\documentclass{article}
\usepackage{PRIMEarxiv}

\usepackage[utf8]{inputenc} 
\usepackage[T1]{fontenc}    
\usepackage{hyperref}       
\usepackage{url}            
\usepackage{booktabs}       
\usepackage{amsmath,amssymb,amsfonts}       
\usepackage{nicefrac}       
\usepackage{microtype}      
\usepackage{fancyhdr}       
\usepackage{graphicx}       
\graphicspath{{media/}}     

\usepackage{cite}
\usepackage{textcomp}
\usepackage{subcaption}
\usepackage[flushleft]{threeparttable}
\usepackage{multirow}
\usepackage{algorithmic, algorithm}
\floatname{algorithm}{Procedure}

\newcommand\algorithmicinput{\textbf{Input:}}
\newcommand\algorithmicoutput{\textbf{Output:}}
\newcommand\INPUT{\item[\algorithmicinput]\item[]}%
\newcommand\OUTPUT{\item[\algorithmicoutput]\item[]}%

\hypersetup{
    colorlinks=true,
    linkcolor=black,
    citecolor=black,
    urlcolor=black
    }

\title{Constructing a meta-learner for unsupervised anomaly detection
\thanks{This work was conducted with the financial support of the Science Foundation Ireland Centre for Research Training in Artificial Intelligence under Grant No. 18/CRT/6223.}
}

\author{
  Małgorzata Gutowska, Suzanne Little, Andrew McCarren\\
  School of Computing, Dublin City University \\
  Dublin, Ireland \\
  \texttt{malgorzata.gutowska2@mail.dcu.ie}, \texttt{\{suzanne.little, andrew.mccarren\}@dcu.ie} 
}

\begin{document}

\maketitle

\begin{abstract}
Unsupervised anomaly detection (AD) is critical for a wide range of practical applications, from network security to health and medical tools. Due to the diversity of problems, no single algorithm has been found to be superior for all AD tasks. Choosing an algorithm, otherwise known as the \textit{Algorithm Selection Problem} (ASP), has been extensively examined in supervised classification problems, through the use of meta-learning and AutoML, however, it has received little attention in unsupervised AD tasks. 
This research proposes a new meta-learning approach that identifies an appropriate unsupervised AD algorithm given a set of meta-features generated from the unlabelled input dataset. The performance of the proposed meta-learner is superior to the current state of the art solution. In addition, a mixed model statistical analysis has been conducted to examine the impact of the meta-learner components: the meta-model, meta-features, and the base set of AD algorithms, on the overall performance of the meta-learner. The analysis was conducted using more than 10,000 datasets, which is significantly larger than previous studies.
Results indicate that a relatively small number of meta-features can be used to identify an appropriate AD algorithm, but the choice of a meta-model in the meta-learner has a considerable impact. 
\end{abstract}

\vspace{20pt}

\input{paper/1_intro}
\input{paper/2_related_work}

\input{paper/3_methodology}
\input{paper/4_results}
\input{paper/5_discussion}
\input{paper/6_conclusions}

\bibliographystyle{unsrt}  
\bibliography{references}

\end{document}

%% file: paper/1_intro.tex
\section{Introduction}
\label{sec:introduction}

Anomaly detection (AD) is gaining increased attention in various business sectors, primarily due to the growing number of systems that collect and use data generated through a variety of daily activities. Detecting anomalies in sophisticated systems not only helps them run smoothly, but in many domains, an identified anomaly is of genuine value. An outlier can be a potentially harmful event that should be avoided or an early indicator of a new trend. Insurance, finance, network traffic monitoring, and health are just a few examples of the many different fields that benefit from AD applications~\cite{Campos2016, surveyAD2019ieee, surveyDLAD2019arxiv, 2021unifying_review}.

In contrast to typical classification tasks, most real-world anomaly detection problems are unsupervised. Anomalies are rarely known in advance since they have not yet been identified, and those that could have been detected may differ significantly from those still to be discovered. This is intuitive given the essential properties of anomalies, as they are markedly different from other data instances and rare compared to normal data points~\cite{Goldstein2016}.

To overcome the challenges posed by the difficult to define character of anomalies, researchers categorise them in a variety of ways, such as by their relationship to other data points~\cite{Chandola_2009}:
\begin{itemize}
\item point anomaly -- an individual data instance deviating from other data, often referred to as global anomaly~\cite{schubert2014local, Goldstein2016, wang2018randomwalk},
\item collective anomaly -- a collection of data instances anomalous with respect to the entire dataset,
\item contextual anomaly -- a data instance anomalous only in a specific context but not otherwise, also referred to as a local anomaly~\cite{wang2018randomwalk}.
\end{itemize}

Typically, AD algorithms seek to estimate the likelihood of every data point being an anomaly in an unsupervised manner~\cite{Goldstein2016}, considering the anomaly characteristics and types mentioned above.
These algorithms return an anomaly score (usually not normalised), where a higher score indicates an increased likelihood that a given data point is an anomaly~\cite{Chandola_2009,Goldstein2016}. It is not possible, therefore, to measure an algorithm's performance in the form of a single metric such as accuracy, recall or precision. A detailed study of the results is usually necessary to assess the performance for a new task.

\subsection{Algorithm Selection Problem}
Although numerous AD techniques have been proposed, it is generally accepted that no single method is optimal for all AD problems~\cite{Campos2016,surveyAD2019ieee}.
The concern of a single algorithm performing well on a limited number of problems is not unique to AD. Rather, it is common across machine learning tasks, and is otherwise known as the \textit{No Free Lunch Theorem}~\cite{wolpert1997nofreelunch}. The problem of selecting the best algorithm for a given task is referred to as \textit{Algorithm Selection Problem} (ASP)~\cite{ali2017asp, bischl2016aslib, khan2020metalearningsurv}.
Conventional approaches to the ASP are not always optimal. For example, the trial-and-error strategy can easily become time-consuming, and hiring a human expert in a specific field can be prohibitively expensive. Furthermore, none of these strategies guarantees success.

An alternative method to overcome these drawbacks is the automated selection of the best performing algorithm. Creating a model that learns from historical evaluations is a well-established approach to ASP, also known as meta-learning~\cite{metalearning_surv2015lemke,bookAutoML2019}. Meta-learning, along with Automated Machine Learning (AutoML), of which meta-learning is an essential part, is a rapidly developing topic in machine learning~\cite{bookAutoML2019}.

\begin{figure}
    \centering
    \includegraphics[width=0.6\columnwidth]{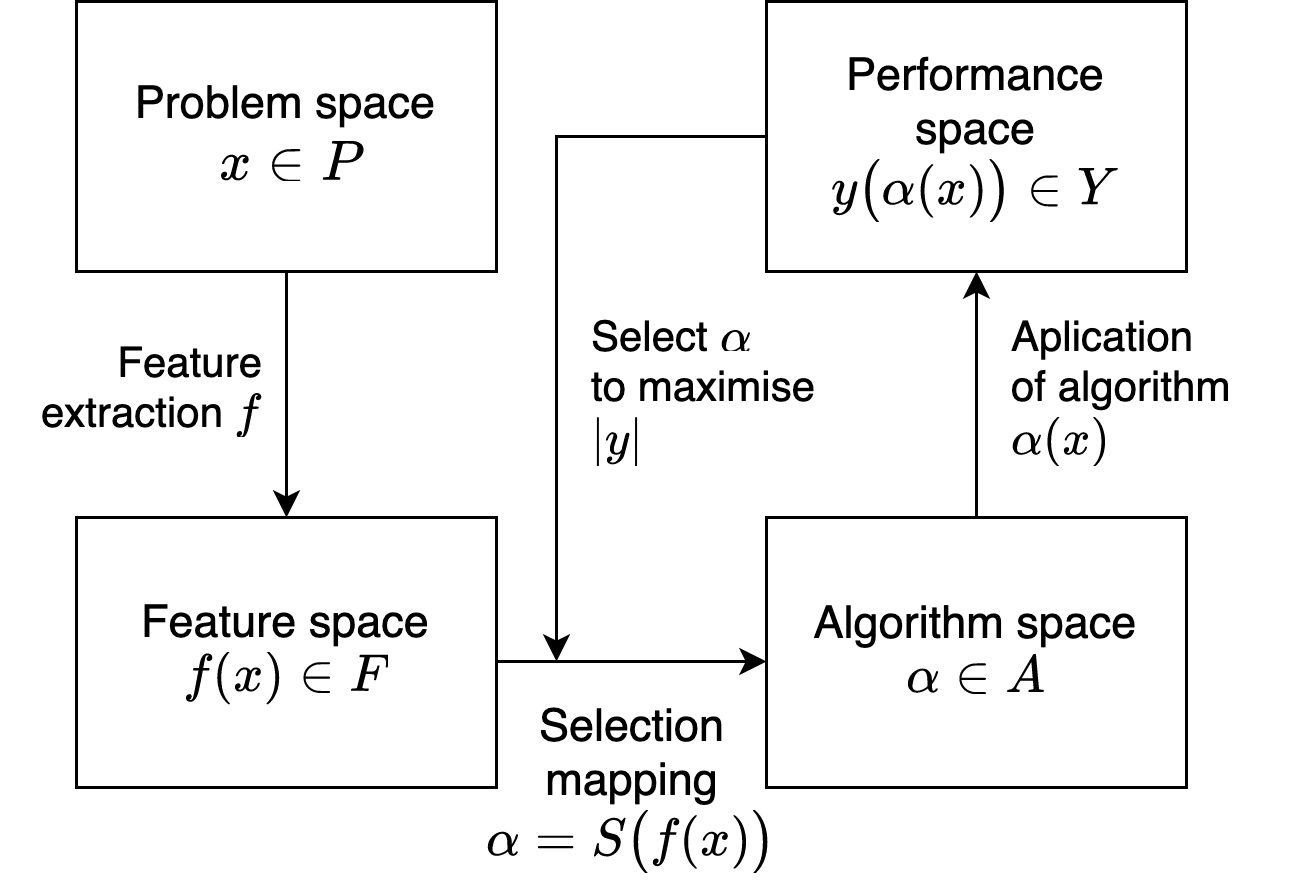}
\caption{Rice's framework, as presented by Smith-Miles~\cite{smithmiles2009metalearning_surv}.}
\label{fig:rice_framework}
\end{figure}

Among the most popular strategies for meta-learning is the use of dataset characteristics that can be mapped to the performance of specific algorithm configurations from historical evaluations~\cite{bookAutoML2019, smithmiles2009metalearning_surv}.
Rice's seminal paper~\cite{rice1976algorithm} conceptualised the problem of algorithm selection by mapping problem characteristics to algorithm performance. This framework was expanded by Smith-Miles~\cite{smithmiles2009metalearning_surv} and captured in the form shown in Fig.~\ref{fig:rice_framework}.

There are four essential components of the framework~\cite{smithmiles2009metalearning_surv}:
\begin{itemize}
\item problem space $P$ -- a set of dataset instances of a problem,
\item feature space $F$ -- a set of characteristics generated from each dataset instance $x$,
\item algorithm space $A$ -- a set of algorithms (possibly including variations incorporating hyperparameters),
\item performance space $Y$ -- a set of performance metrics of the algorithms from $A$ evaluated over the problem space $P$.
\end{itemize}
The ASP can be defined as 
``For a given problem instance $x \in P$ with features $f(x) \in F$, find the selection mapping $S(f(x))$ into algorithm space $A$, such that the selected algorithm $\alpha \in A$ maximises the performance mapping $y(\alpha (x)) \in Y$''~\cite{smithmiles2009metalearning_surv}.

The ASP is even more challenging when faced with an unsupervised task. Applying a trial-and-error approach or maximisation of performance mapping is problematic without a single performance metric. A meta-learner that indicates the best algorithm for a new unsupervised task is thus a necessary tool, particularly in scenarios with high up-front uncertainty and high variability.

Few studies have proposed an automated algorithm selection in AD~\cite{kandanaarachchi2019instance, metaod2021, ad_metafeatures2021, uod2021ensemble, n1experts2022}. Of these, only one model is designed for unsupervised scenarios and utilises the meta-learner framework as presented by Rice~\cite{metaod2021}.

The current work has been motivated by the growing importance of AD tools and the little research done to address the problem of algorithm selection in unsupervised AD scenarios. In this research, an alternative meta-learner for unsupervised AD has been developed with respect to its three components: meta-feature space $F$, AD algorithm space $A$, and the meta-model (i.e., a model that learns the mapping of the meta-features into the algorithm space~\cite{metalearning2019chapter}, $S(f(x))$, referred to as $m$ in the following part of this manuscript). A comparison with the state of the art and a statistical analysis of eight meta-learner variants (three factors, each with two levels) was then performed.

This study makes several contributions to the algorithm selection problem in unsupervised anomaly detection. The main ones include:
\begin{itemize}
\item creation of an alternative unsupervised AD meta-learner, superior to the existing state of the art solution,
\item examining the effect size of the components contributing to the performance of the meta-learner.
\end{itemize}
In addition, this study presents secondary contributions:
\begin{itemize}
\item proposal of an extension of Rice's framework, which is refactored to the unsupervised AD problem,
\item implementation of an existing state of the art unsupervised meta-learning technique on an AD benchmark project consisting of more than 10,000 diverse datasets.
\end{itemize}

The remainder of the paper is structured as follows: Section~\ref{sec:related-work} provides an overview of work in the meta-learning and AutoML domains, as well as work in the algorithm selection problem for unsupervised anomaly detection. The conducted experiments are described in Section~\ref{sec:methodology}, and the results are reported and discussed in Sections~\ref{sec:results} and~\ref{sec:discussion}, respectively. The study's conclusions are presented in Section~\ref{sec:conclusions}.

%% file: paper/2_related_work.tex
\section{Related work} \label{sec:related-work}

Since the formalisation of the ASP framework by Rice, meta-learning has received increased attention in the research community. Studies have focused on specific areas of meta-learning, such as hyperparameter optimisation (HO)~\cite{komer2014hyperopt, bayesian_hyperparam_opt2015, tuneornot2015, hyperparam_opt2016, Sanders2017hyperopt, wistuba2016hyperopt, hypertuningsvm2019, Mantovani2020hyperopt} or meta-features development~\cite{metafeatures_general2016, ABDRASHITOVA2018metafeatures, gutierrez2019metafeatures}. Several survey studies have also presented a large-scale overview of the research~\cite{metalearning_surv2002, smithmiles2009metalearning_surv, metalearning_survey2015munoz, metalearning_surv2015lemke, khan2020metalearningsurv}. In particular, a comprehensive compilation of the research and achievements in meta-learning and broader fields such as AutoML is presented in~\cite{bookAutoML2019}. Current AutoML systems are not only capable of performing model or hyperparameter selection tasks, but are also fully functional automated pipelines of processes that include training and testing without the need for human intervention~\cite{guyon2019analysis}.
To date, several automated AutoML systems have been developed, including Auto-WEKA~\cite{autweka2019}, Auto-Sklearn~\cite{autosklearn2015}, Auto-Sklearn 2.0~\cite{autosklearn2_2020}, Hyperopt-Sklearn~\cite{komer2014hyperopt}, Auto-Net~\cite{autonet2019} and others~\cite{olson2016tpot, steinruecken2019automatic}.

These systems were developed to solve supervised problems. The ASP for unsupervised settings remains mostly unexplored. A few studies have examined the meta-features' potential in AD scenarios~\cite{Campos2016, 2020Kandanaarachchi_meta,metaod2021,ad_metafeatures2021}, but their limitations are either due to the required knowledge of data labels or a structure, or a need for an extensive evaluation of algorithms for a new task. The approach presented in~\cite{metaod2021} has addressed the ASP problem for unsupervised tasks by utilising the knowledge gained through past algorithm evaluation and is the state of the art solution for the aforementioned problem.

A key early exploration of this challenge at scale by Campos \textit{et al.}~\cite{Campos2016} examined how well the selected AD methods handled anomalies across datasets. The authors defined two dataset properties, which were \textit{difficulty} and \textit{diversity}. The difficulty score indicated, for each AD method, the difficulty of detecting outliers in a given dataset. The diversity score reflected the agreement between the methods on the difficulty score. The researchers created a feature map based on these two scores and positioned the datasets accordingly. This study explored meta-analysis of the algorithm's performance across datasets, and is viewed as a first step toward inferring dataset meta-features and developing a system to aid in algorithm selection.

Kandanaarachchi \textit{et al.}~\cite{2020Kandanaarachchi_meta} attempted to create a method for automated algorithm selection in AD problems by analysing the problem space (referred to as instance space). Two factors that could impact algorithm performance were investigated. These were dataset normalisation and dataset characteristics (meta-features). Two meta-features derived from an initial set of 362 features were used to characterise the datasets and create a two-dimensional space filled with the dataset instances. The feature space outlined optimal regions for a specific method and normalisation type. This study examined 12,000 datasets obtained by downsampling and repurposing classification datasets, yielding the most extensive published set of AD benchmark datasets to date.
However, the meta-features developed in this work required labelled data to be present, making the chosen features unsuitable for an unsupervised AD problem.

An alternative set of meta-features for AD tasks has been proposed by Kotlar \textit{et al.}~\cite{ad_metafeatures2021}. Their research focused on semantic features designed explicitly for AD, such as global, local, or collective anomaly types, as well as other characteristics like \textit{anomaly space}, \textit{anomaly ratio}, \textit{data type}, or \textit{data domain}. Creating these features requires prior knowledge of the anomalous instances, such as their ratio or distribution. These features also need to be provided by a human expert rather than extracted automatically from the datasets. This work cannot, therefore, be directly applied to an automated unsupervised meta-learning problem.

Recent work from Zhao \textit{et al.}~\cite{metaod2021} has provided a system (referred to as \textit{MetaOD} or \textit{UOMS}: \textit{Unsupervised Outlier Model Selection}) for predicting the best algorithm for an unsupervised AD task. This work utilised inherent dataset characteristics that do not require labelled data and are generated automatically from datasets. In addition to the \textit{statistical meta-features} common to many classification tasks, \textit{landmarking features} specific to AD problems were also included. A collection of algorithms are assessed covering classic AD algorithms, each combined with a set of hyperparameters to yield 302 potential AD models. The experiments were conducted on two test setups with 100 and 63 datasets, respectively.
The meta-model was specifically designed to address the algorithm selection problem in AD and is based on collaborative filtering (CF), employing the matrix factorisation method. This approach is further highlighted in Section~\ref{sec:methodology}, and, as a state of the art solution, it forms a key part of the comparative analysis in the current study.

A different approach of addressing the ASP in AD to that mentioned above has recently been presented in~\cite{uod2021ensemble, n1experts2022}.
These studies consider the unsupervised nature of AD problems but do not use offline meta-training. Instead, the systems for assessing the performance of a predefined set of algorithms ``online'', while evaluating them on a new task, were proposed. The primary shortcoming of these systems is non-scalability. They require evaluation (or even multiple evaluations) of a number of algorithms when confronted with a new problem, without taking advantage of transfer learning from historically evaluated algorithms.

Although ASP and AutoML have received a lot of interest overall, limited research has examined the ASP in unsupervised AD, i.e. when labels for a task at hand are unknown. To date, the most successful solution presented in the research is the \textit{UOMS} outlined by~\cite{metaod2021}. The lack of practical benchmarks or potential directions for the improvement of future systems is another obstacle. In most cases, the trend in ASP for AD problems tends to focus on the development of meta-features and not the meta-model or the range of applicable algorithms that can be chosen by the meta-learner. Understanding the influencing factors in a meta-learner is key for the future development of meta-learning systems. In addition, little attention has been given to creating an extensive dataset that can truly test a proposed meta-learner.

%% file: paper/3_methodology.tex
\section{Methodology} \label{sec:methodology}
This section outlines an alternative strategy to that proposed by Zhao \textit{et al.}~\cite{metaod2021}, which uses fewer meta-features and has a neural network (NN) as opposed to CF as the base learner. An extensive dataset is required to validate this approach, and the process for generating this data is initially proposed. An evaluation process is then outlined, which attempts to identify the components that contribute to the identification of the best performing algorithm when using meta-learning for a range of unsupervised AD algorithms. 

\subsection{Data generation}
The problem/dataset space $P$ for this experiment was built on one of the largest available unsupervised AD dataset repositories. A total of $N=10,000$ datasets were randomly chosen from those proposed in~\cite{2020Kandanaarachchi_meta} \footnote{\url{https://bridges.monash.edu/articles/dataset/Datasets_12338_zip/7705127/4}}. Each dataset is defined as $x_i$ where $i=1,\dots,N$.

To generate the performance space $Y$, a selection of AD algorithms $\alpha_j \in A$ were evaluated over all the datasets in $P$. During the algorithm implementation, two performance metrics were captured: the area under the receiver operating characteristic curve (AUC) and the average precision (AP).

The AUC and the AP are the most widely used evaluation measures in anomaly detection~\cite{Campos2016, surveyAD2019ieee}.
The output of an AD method is typically an anomaly score, which indicates the likelihood that the observation is anomalous. 
The translation of anomaly scores into binary labels that are required in common performance metrics, such as accuracy or precision, involves using a threshold, which in many practical AD cases is unknown without additional data exploration. The AUC and AP measures provide the integral result for all thresholds from 0 to 1, making them threshold-independent.

The result of this experiment stage were two matrices of the performance values $\textbf{Y}_{\nu} \in \mathbb{R}^{N \times L}$, where $\nu \in \left\{\text{AUC}, \text{AP} \right\}$, whereas $N$ and $L$ denote the number of the datasets and AD algorithms, respectively.

Procedure~\ref{alg:data-generation} outlines the steps taken for generating the data of the performance values $\textbf{Y}_{\text{AUC}}$ and $\textbf{Y}_{\text{AP}}$ for the AD algorithm set $A$ and the set of datasets $P$.

{\centering
\begin{minipage}{.6\linewidth}
\begin{algorithm}[H] \caption{AD algorithms performance data generation} \label{alg:data-generation}
\begin{algorithmic}[1] 
{\fontsize{9}{13}\selectfont
\INPUT
\hskip\algorithmicindent $A = \left\{\alpha_j \,|\; j=1,\dots, L \right\}$ \\
\hskip\algorithmicindent $P = \left\{x_i \,|\; i=1,\dots, N \right\}$ \\

\OUTPUT
\hskip\algorithmicindent $\textbf{Y}_{\texttt{AUC}} \in \mathbb{R}^{N \times L}$ \\
\hskip\algorithmicindent $\textbf{Y}_{\texttt{AP}} \in \mathbb{R}^{N \times L}$ \\

\FOR{{$j=1$ to $L$}}
\FOR{$i=1$ to $N$}
\STATE[Run algorithm $\alpha_j$ over dataset $x_i$] \(
\left. \begin{array}{l}
y_{\texttt{AUC}\>ij} \\
y_{\texttt{AP}\>ij}
\end{array} \right\}\leftarrow  \alpha_j(x_i)
\)
\ENDFOR
\ENDFOR
\RETURN $\textbf{Y}_{\texttt{AUC}}$, $\textbf{Y}_{\texttt{AP}}$
}
\end{algorithmic} 
\end{algorithm}
\end{minipage}
\par
}

The details of the AD algorithms used are discussed in Section~\ref{sec:alt-meta}. The algorithm performance metrics obtained at this stage are available at http://ieee-dataport.org/10491.

\subsection{Proposed Meta-Learner} \label{sec:alt-meta}
The meta-learner proposed in the current study has three component parts: meta-features ($F_1$), a set of base AD algorithms ($A_1$), and the base learner -- meta-model ($m_p$). Each component is outlined below.

\subsubsection{Meta-features}
The current study uses a significantly smaller set of meta-features ($F_1$) compared to UOMS~\cite{metaod2021} (19 vs 200), which were specifically designed to accommodate a broad range of anomaly characteristics.
The overall design criteria for the features are as follows:
\begin{itemize}
\item to be independent of the data labels (suitable for unsupervised scenarios),
\item to capture the main types of anomalies, such as global, local, and collective,
\item to describe multivariate characteristics of the data (mutual relations between the data points and their neighbourhood).
\end{itemize}

There are two steps involved in the process of feature generation: 1) calculation of distances between data points; 2) obtaining the actual features as statistical characteristics of the acquired distance distributions. 

In step 1, an approach inspired by Moran scatterplots depicting the relationship between global and local z-scores~\cite{schubert2014local} has been proposed to express characteristics related to local and global anomalies.
The approach has been generalised to multivariate datasets by substituting the Mahalanobis distance for z-scores.
Global and local Mahalanobis distances, $\zeta_G^i$ and $\zeta_{Ln}^i$, were calculated for each data point $i$ as expressed in~(\ref{eq:mah-g}) and~(\ref{eq:mah-l}):
\begin{equation}
    \zeta_G^i (\overrightarrow{z_i}) = \sqrt{( \overrightarrow{z_i} - \overrightarrow{ \mu })^T \mathbf{S}^{-1} ( \overrightarrow{z_i} - \overrightarrow{ \mu })} 
    \label{eq:mah-g}
\end{equation}

\begin{equation}
    \zeta_{Ls}^i (\overrightarrow{z_i}) = \sqrt{( \overrightarrow{z_i} - \overrightarrow{ \mu_{s} })^T \mathbf{S_s}^{-1} ( \overrightarrow{z_i} - \overrightarrow{ \mu_{s} })}
    \label{eq:mah-l}
\end{equation}
where \(\overrightarrow{z_i} = (z_{1i}, ..., z_{Ki})\) represents a single observation $i$ (a data point) with \(K\) features, \(\overrightarrow{\mu} = (\mu_1, \mu_2, ..., \mu_K)\) represents the mean of all other observations in the dataset, and \(\mathbf{S}\) is the covariance matrix. In (\ref{eq:mah-l}), \(\overrightarrow{\mu_s}\) and \(\mathbf{S_s}\) represent the mean and covariance matrix of the \(s\) nearest neighbours of \(\overrightarrow{z_i}\), and \(\zeta_{Ls}^i\) defines the Mahalanobis distance to the respective $s$ nearest neighbours. The number of nearest neighbours has been chosen using a grid search approach and defined as $s \in\{20, 60, 80\}$. After this step, each data point $i$ has been represented by four distances to the rest of the data: $\left \{\zeta_G^i, \zeta_{L20}^i,  \zeta_{L60}^i, \zeta_{L80}^i \right \}$. 

In step 2, the statistical characteristics of each of the distance profiles have been obtained. 
For each set of \(\zeta_{Ls}\) and \(\zeta_G\) the features such as \textsc{TotalRange} $\textsc{TR}$, \textsc{CenterMass} $\textsc{CM}$, \textsc{TailHalf} $\textsc{TH}$, and \textsc{TailQuarter} $\textsc{TQ}$ have been calculated as expressed in~(\ref{eq:tr})--(\ref{eq:tq}):
\begin{equation} \label{eq:tr}
	\textsc{TR}_{\zeta*} = \max(\zeta_*^i) - \min(\zeta_*^i)
\end{equation}
\begin{equation}
	\textsc{CM}_{\zeta*} = \frac{1}{\textsc{TR}} \left(P^{75}(\zeta_*^i) - P^{25}(\zeta_*^i)\right)
\end{equation}
\begin{equation}
	\textsc{TH}_{\zeta*} = \frac{1}{\textsc{TR}} \left( \max(\zeta_*^i) - P^{50}(\zeta_*^i) \right)
\end{equation}
\begin{equation} \label{eq:tq}
	\textsc{TQ}_{\zeta*} = \frac{1}{\textsc{TR}} \left(\max(\zeta_*^i) - P^{75}(\zeta_*^i)\right)
\end{equation}
where \(\zeta_*\) is one of \(\zeta_G\), \(\zeta_{L20}\), \(\zeta_{L60}\), \(\zeta_{L80}\) and \(P^{25}\), \(P^{75}\), and \(P^{50}\) are the 25\textsuperscript{th} and 75\textsuperscript{th} percentiles and the median, respectively.
In addition, for each neighbourhood $s$ a property aiming to describe a dataset ``locality'' \(\mathcal{L}_s\) was calculated using~(\ref{eq:locality}):
\begin{equation} \label{eq:locality}
	\mathcal{L}_s = \frac{1}{N}\sum_{i}^{N} \frac{\zeta_{Ls}^{i}}{\zeta_{G}^i}
\end{equation}
with \(N\) representing the total number of data instances within the dataset. The resulting features are summarised in Table~\ref{tab:meta-features}.

\begin{table}
\caption{Meta-features proposed in the $F_1$ set.}
\centering
\begin{tabular}{llc}
\toprule
Meta-feature & Instances & Count \\ \midrule
$\textsc{TR}_{\zeta*}$ & \multirow{4}{*}{$\zeta_*$: \{$\zeta_G$, $\zeta_{L20}$, $\zeta_{L60}$, $\zeta_{L80}$\}} & 4 \\
$\textsc{CM}_{\zeta*}$ & & 4  \\
$\textsc{TH}_{\zeta*}$ & & 4 \\
$\textsc{TQ}_{\zeta*}$ & & 4 \\
$\mathcal{L}_s$ & $s$: \{20, 60, 80\} & 3 \\ \bottomrule
\end{tabular}
\label{tab:meta-features}
\end{table}

\subsubsection{Set of anomaly detection algorithms}
The requirement to select methods from a few diverse families, in particular, several methods from the ``classic bucket'', and modern deep learning-based techniques, drove the selection for the base set of meta-learner algorithms $A_1$. The set includes ten ``classic'' and three neural network-based AD algorithms. Table~\ref{tab:small-model-set} presents the algorithms and their parameters chosen for $A_1$.
In contrast to UOMS~\cite{metaod2021}, this collection did not focus on examining the algorithm hyperparameters. This is therefore referred to as ``w/o HO''. The lack of the HO makes this set more distinct to UOMS that is employed for the comparative evaluation.

\begin{table}[ht]
\caption{AD models and their parameters comprising the set proposed in this study ($A_1$).}
\label{tab:small-model-set}
\centering
\begin{tabular}{lll}
\toprule
AD algorithm & Parameter 1 & Parameter 2 \\ \midrule
LOF & n\_neighbors = 60 & distance = 'euclidean' \\
kNN & n\_neighbors = 60 & method = 'mean' \\
k\textsuperscript{th}NN & n\_neighbors = 60 & method = 'largest' \\
OCSVM & nu = 0.008 & kernel = 'rbf' \\
COF & n\_neighbors = 60 & N/A \\
ABOD & n\_neighbors = 60 & N/A \\
iForest & n\_estimators = 100 & max\_features = 1.0 \\
HBOS & n\_bins = 90 & tolerance = 0.5 \\
COPOD & N/A & N/A \\
PCA & n\_components = 'mle' & svd\_solver = 'full' \\
VAE & epochs = 500 & hidden layers, as described in~(\ref{eq:vae-hidden-layers}) \\
SO-GAAL & epochs = 25 & N/A \\
MO-GAAL & epochs = 25 & N/A \\ \bottomrule
\end{tabular}
\end{table}

The conventional methods involved in the experiments cover the following AD algorithms: Local Outlier Factor (LOF)~\cite{lof2000}, k-Nearest Neighbours (kNN)~\cite{ramaswamy2000knn}, One-Class Support Vector Machines (OCSVM)~\cite{scholkopf2000support}, Connectivity-Based Outlier Factor (COF)~\cite{tang2002cof}, Angle-based Outlier Detector (ABOD)~\cite{kriegel2008abod}, Isolation Forest (iForest)~\cite{isolationforest2008}, Histogram-based Outlier Score (HBOS)~\cite{hbos}, Lightweight Online Detector of Anomalies (LODA)~\cite{pevny2016loda}, Copula-based Outlier Detection (COPOD)~\cite{li2020copod}, and Principal Component Analysis-based anomaly detection (PCA)~\cite{shyu2003pcaanomaly}. 
Deep learning-based approaches include Variational Autoencoder (VAE)~\cite{kingma2013vae}, Single-Objective Generative Adversarial Active Learning (SO-GAAL), and Multi-Objective Generative Adversarial Active Learning (MO-GAAL)~\cite{liu2019gaal}.

The VAE's architecture was designed individually for each dataset. The number and dimensions of the hidden layers were determined by the number of features in each dataset. Given that $K$ represents the number of features in a given dataset, the hidden layers were built as described in~(\ref{eq:vae-hidden-layers}).
\begin{equation}
\label{eq:vae-hidden-layers}
	\left\{ \begin{array}{ll}
    K \times [0.75, 0.5, 0.33, 0.25] & \text{if} \quad K \geq 100 \\
    K \times [0.75, 0.5, 0.25] & \text{if} \quad 100 > K \geq 50 \\
    K \times [0.5, 0.25] & \text{if} \quad 50 > K \geq 6 \\
    2 & \text{if} \quad K < 6
    \end{array} \right.
\end{equation}

The implementation from the Python PyOD package~\cite{zhao2019pyod} was used to perform the evaluation of the algorithms.

\subsubsection{Meta-model}
A key part of a meta-learner is a meta-model, whose goal is to select the AD algorithm that performs best on an unseen and unlabelled AD dataset.
The meta-model proposed in this study ($m_1$) is based on a neural network architecture. Neural networks have proven successful in a variety of tasks and are relatively time-efficient when training for small-size problems.

The architecture proposed in this work features three hidden layers (64, 64, and 32 nodes), a dropout of 0.2 after each layer, and the predictive regression layer that outputs the predicted performance values $\hat{y}$ for each  algorithm $\alpha_1,\dots, \alpha_L$, with $L$ being the number of algorithms used in the meta-training. 

\begin{figure}
    \centering
    \includegraphics[width=0.5\linewidth]{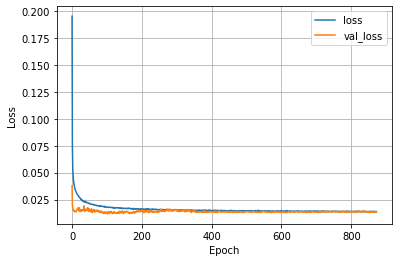}
\caption{Training loss (\textit{loss}) and validation loss (\textit{val\_loss}) versus epochs in training one of the NN-based meta-model variant.}
\label{fig:loss}
\end{figure}

The training of each meta-model has been started from random weights and run through up to 1000 epochs. The \textit{Early Stopping} functionality from Keras library~\cite{chollet2015keras} has been implemented to cease further training when no improvement is observed, as measured by the loss on the validation data. Subsequently, the optimal weights have been obtained based on the validation loss. The resulting training length ranged for different meta-learner variants (variants described in Section~\ref{sec:eval-strat}) from 300 to 1000 epochs. Fig.~\ref{fig:loss} presents an example of the training process picturing the training and validation loss in subsequent epochs. The batches of 32 samples were used in training and the Adam optimiser was employed to minimise the mean squared error.

The network architecture parameters, such as the number of hidden layers, the number of nodes, dropout level, epoch count, and batch size, were chosen using the grid search approach and optimised with the \textit{Weights and Biases} tool~\cite{wandb}. The search for the optimal architecture ranged from 2-layered networks of 4 + 4 nodes to 3-layered networks of 64 + 64 + 64 nodes. More complex architectures were not considered to avoid the risk of model overfitting.

\subsection{Evaluation Strategy} \label{sec:eval-strat}
\begin{figure*}
    \centering
    \includegraphics[width=\textwidth]{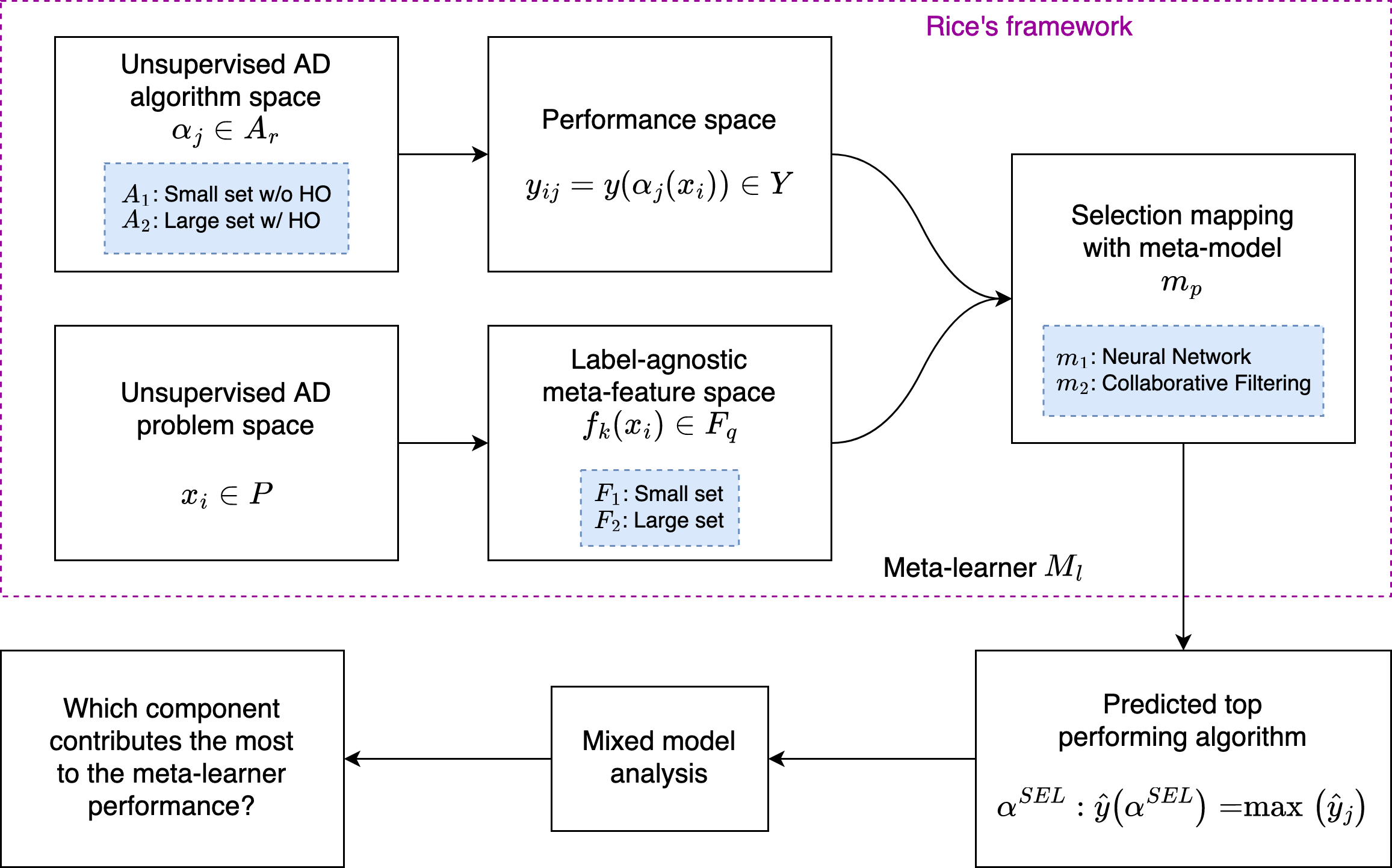}
\caption{Experimental design including the meta-learner framework with reference to Rice's representation (outlined with a purple line) and factor components used in comparative evaluation (blue stickers). $A_1$, $F_1$, $m_1$ and $A_2$, $F_2$, $m_2$ relate and refer to components proposed in the current study and UOMS, respectively.}
\label{fig:exp}
\end{figure*}

A $2^3$-factorial design was implemented to examine the factors which contributed to the performance of meta-learners for unsupervised AD. The factors used in this experiment (meta-feature generation strategies $F_q$, the base set of AD algorithms $A_r$, and the meta-model $m_p$) were from the approach proposed in the current study and the UOMS~\cite{metaod2021}. Table~\ref{tab:learners-summary} summarises the factors of both approaches. The experimental design is illustrated in Fig.~\ref{fig:exp}.

\begin{table}
\centering
\caption{Summary of the factor components of meta-learners compared in this study.}
\label{tab:learners-summary}
\begin{tabular}{p{0.21\linewidth}|p{0.2\linewidth}p{0.35\linewidth}}
\toprule
 & Current study & UOMS \\ \midrule
Meta-model $m_p$ & NN & CF \\ \midrule
Meta-features $F_q$ & 19 features specific to AD problems & 200 features, combined: statistical and landmarking features \\ \midrule
Set of AD algorithms $A_r$ & 13 algorithms w/o HO & 298 models: 8 algorithms combined with sets of hyperparameters \\ \bottomrule
\end{tabular}
\end{table}

To perform the experiment, eight meta-learners \(M_l := M_{pqr}\), were designed, where $l = 1,\dots, 8$, and $p,q,r \in \left\{1,2 \right\}$, which incorporated two variants of mentioned factors $F_q$, $A_r$, and $m_p$. Indices \textit{1} and \textit{2} have been used to denote factors from the current study and UOMS, respectively. Each combination of the $2^3$-factorial design was implemented on each candidate dataset $x_i$ producing the predicted performance metrics $\hat{y}_{ij}$ for each algorithm $\alpha_j$. The performance metrics of meta-learners' selected algorithms $\alpha^{\texttt{SEL}} \in A$ were then analysed using a mixed model analysis, where $x_i$ was considered to be the subject. The algorithm $\alpha^\texttt{SEL}$ was selected from $\alpha_j \in A$ for each $x_i$ to maximise the predicted performance $\hat{y}_{ij}$. The steps performed are described in Procedure~\ref{alg:metamodel-training} and~\ref{alg:select-top}, whereas the functional form for this approach is shown in~(\ref{eq:meta-model}) and~(\ref{eq:alg-selection}). For each $p,q,r \in \left\{ 1,2 \right\}$:

\begin{equation}
    \hat{y}_{ij} = \hat{y}_i(\alpha_j) = m\big(F(x_i)\big)
    \label{eq:meta-model}
\end{equation}

\begin{equation}
    y_i^\texttt{SEL} = y_{i}(\alpha^\texttt{SEL}) : \hat{y_i}(\alpha^\texttt{SEL}) := \max(\hat{y}_{ij})
    \label{eq:alg-selection}
\end{equation}
where
\[\left\{ \begin{array}{lll}
m & = & m_p \\
F & = & F_q \\
\alpha_j & \in & A_r
\end{array} \right.\]
The indices $p,q,r$ have been omitted in~(\ref{eq:meta-model}) and~(\ref{eq:alg-selection}) for readability purposes.

{\centering
\begin{minipage}{.6\linewidth}
\begin{algorithm}[H] \caption{Training of meta-models} \label{alg:metamodel-training}
\begin{algorithmic}[1]

{\fontsize{9}{13}\selectfont
\INPUT 
$\textbf{F}_q \in \mathbb{R}^{N \times K_q}$ \\
$\textbf{Y}_{r\, \texttt{AUC}}$, $\textbf{Y}_{r\, \texttt{AP}} \in \mathbb{R}^{N \times L_r}$, where $\textbf{Y}_r = Y(A_r)$ \\

\OUTPUT
$Y_{\texttt{AUC}}^{\texttt{SEL}}$, $Y_{\texttt{AP}}^{\texttt{SEL}} \in \mathbb{R}^{N^{\texttt{test}}}$ \\

\FORALL{$\nu \in \left\{\texttt{AUC}, \texttt{AP} \right\}$}
\FORALL{$F_q : q \in \left\{1,2\right\}$}
\FORALL{$Y_r : r \in \left\{1,2\right\}$}
\FORALL{$m_p : p \in \left\{1,2\right\}$}
\STATE[Split the input and output data]\\ 
$(F_q^{\texttt{train}}, F_q^{\texttt{test}}) \leftarrow F_q$
\STATE $(Y_r^{\texttt{train}}, Y_r^{\texttt{test}}) \leftarrow Y_r$
\STATE[Train the meta-model in a supervised manner] \\
$m_p \leftarrow \text{train}(F_q^{\texttt{train}}, Y_r^{\texttt{train}})$
\STATE[Predict] $\widehat{\mathbf{Y}}_r^{\texttt{test}} \leftarrow m_p(F_q^{\texttt{test}})$ 
\STATE[Select the best algorithm] Procedure \ref{alg:select-top}
\ENDFOR
\ENDFOR
\ENDFOR
\ENDFOR
\RETURN $Y_{\texttt{AUC}}^{\texttt{SEL}}, Y_{\texttt{AP}}^{\texttt{SEL}}$
}
\end{algorithmic} 
\end{algorithm}
\end{minipage}
\par
}

{\centering
\begin{minipage}{.6\linewidth}
\begin{algorithm}[H] \caption{Find the best predicted algorithm's performance} \label{alg:select-top}
\begin{algorithmic}[1]
{\fontsize{9}{13}\selectfont
\INPUT $\widehat{\textbf{Y}}^{\texttt{test}} \in \mathbb{R}^{N^{\texttt{test}} \times L}$
\OUTPUT $Y^{\texttt{SEL}} \in \mathbb{R}^{N^{\texttt{test}}}$

\FOR{$i=1$ to $N^{\texttt{test}}$}
\STATE $y_i^{\texttt{SEL}} \leftarrow y_i(\alpha^{\texttt{SEL}}) \,:\, \hat{y}_i(\alpha^{\texttt{SEL}}) := \max(\hat{y}_{ij})$
\ENDFOR
\RETURN $Y^{\texttt{SEL}}$
}
\end{algorithmic} 
\end{algorithm}
\end{minipage}
\par
}

\begin{figure*}
    \centering
    \includegraphics[width=0.9\textwidth]{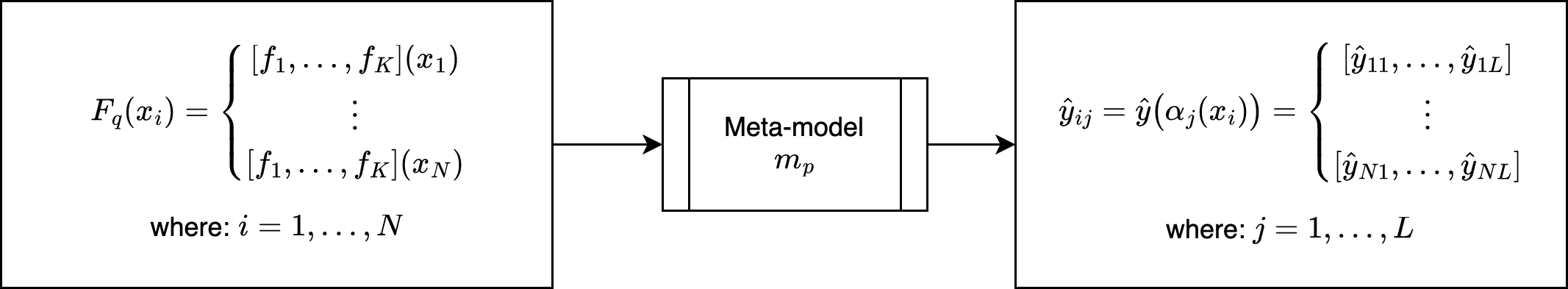}
\caption{Meta-model framework applied and the multi-factor response.}
\label{fig:meta-model}
\end{figure*}

Figure~\ref{fig:meta-model} depicts the meta-model framework, including the multi-factor response, as described in~(\ref{eq:meta-model}).

\subsubsection{Meta-learner training}
The meta-model training in this study was conducted using a typical supervised machine-learning pipeline, where the datasets $x_i$ were split in a 60:15:25 ratio into the train, validation, and test sets. The meta-model of each meta-learner $M_l$ was trained to find a mapping between the relevant meta-features and the AD model performance metrics. The validation set was used to help inform the decision to end the training process.
The learnt mapping was then applied to the new set of test datasets $x_i^{\texttt{test}}$ where the performance values of each AD algorithm $\hat{y}_{ij}$ were predicted.

\subsubsection{Comparison of two meta-learners} \label{sec:meth-compar-direct}
The performance measure of the meta-learner $M_l$ on the dataset $x_i$ has been defined as the actual performance $y_i^\texttt{SEL}$ (either AUC or AP) of the selected algorithm $\alpha^\texttt{SEL}$, as expressed in~(\ref{eq:alg-selection}). The value of such a metric would be accordingly higher as the selection got closer to the actual best performing algorithm, reflecting the meta-learner's performance. Another metric used in this study to evaluate meta-learner performance was the meta-learner error $D_i$, which, for the dataset $x_i$ has been defined as a difference between the best measured performance $y_i^\texttt{TOP}$ and the performance of the algorithm selected by meta-learner $y_i^\texttt{SEL}$ as shown in~(\ref{eq:distance}): 
\begin{equation} \label{eq:distance}
    D_i = y_i^\texttt{TOP} - y_i^\texttt{SEL}.
\end{equation}

A direct comparison of the meta-learning approaches from the current study to UOMS~\cite{metaod2021} has been performed. The mean performance $\overline{y}^\texttt{SEL}$ measured as AUC and AP over the set of test datasets, and the meta-learners' mean error $\overline{D}$ have been compared.
The statistical significance has been measured using the \textit{Paired t-Test} (paired difference test). In addition to statistical significance, the practical significance (effect size) has been assessed using Cohen's $d$~\cite{cohen2013statistical} as outlined in~(\ref{eq:cohen-perf}) and~(\ref{eq:cohen-error}):

\begin{equation} \label{eq:cohen-perf}
    d_y = \frac{\overline{y}^\texttt{SEL}_1 - \overline{y}^\texttt{SEL}_2}{s^*_y}
\end{equation}
\begin{equation} \label{eq:cohen-error}
    d_D = \frac{\overline{D}_1 - \overline{D}_2}{s^*_D},
\end{equation}
where subscripts \textit{1} and \textit{2} indicate approaches from this study and UOMS, respectively, and $s^*_y$ and $s^*_D$ are the pooled standard deviation of \textit{1} and \textit{2} distributions of performance values and errors, respectively.
The use of the effect size was motivated by the large sample sizes. The number of observations in such cases makes the variables appear statistically significant. As a result, practical significance is a more useful statistic to recognise. The strength of an effect can be categorized as follows~\cite{cohen2013statistical}: 
$$\textup{small effect} \leq 0.2 < \textup{medium effect} \leq 0.5 < \textup{large effect}.$$

\subsubsection{Statistical analysis} \label{sec:meth-compar-all}
In this study, the factors $Z_c \in \left\{m_p, F_q, A_r \right\}$, $c=1, \dots,3$, contributing to the error of the meta-learners in choosing the correct algorithm were examined using a mixed model analysis~\cite{mixedmodels2012stroup,mixedmodels2004mcculloh,mixedmodels_r2013}. The error of the meta-learner or the \textit{Distance from the Top} $D_{il}$, has been defined as the difference between the measured performance of the highest performing AD model $y_{i}^\texttt{TOP}$ on the dataset $x_i$ and the selected algorithm performance metric $y_{il}^\texttt{SEL}$ for a specific meta-learner $M_l$ as outlined in~(\ref{eq:error}):

\begin{equation} \label{eq:error}
    D_{il} = \ln \left ( y_{i}^\texttt{TOP} - y_{il}^\texttt{SEL} \right ).
\end{equation}

Thirty principal components $V_{1i}, \dots, V_{ni}$ (which explain 93 \% of the variance) were generated from both sets of meta-features and used as covariates to adjust for variability due to the differences between datasets.

The model used is expressed in~(\ref{eq:mma})--(\ref{eq:error-sigma}):
\begin{equation} \label{eq:mma}
\begin{split}
D_{il} = \beta_{0} + \beta_1 Z_{1} + \beta_2 Z_{2} + \beta_3 Z_{3} + \\ 
+ \beta_{1'} V_{1i} +  \cdots + \beta_{n'} V_{ni} + \gamma_{0i} + \varepsilon_{il}
\end{split}
\end{equation}
with
\begin{equation} \label{eq:random-sigma}
\gamma_{0i} \sim N \left ( 0, \sigma_{\gamma}^2 \right ) 
\end{equation}

\begin{equation} \label{eq:error-sigma}
\varepsilon_{il} \sim N \left (0, \sigma_\varepsilon^2 \right )
\end{equation}
where $\beta_0, \dots, \beta_{n'}$ describe the fixed effects~\cite{mixedmodels2012stroup}, and $\gamma_{0i}$ expresses the random effects' intercepts. Fixed coefficients $\beta_0, \dots, \beta_{3}$ were assessed for significance using the $F$-test with statistical significance set at $p < 0.05$. 
The effect size (Cohen's $d$) of the three components $Z_c$ was calculated using~(\ref{eq:cohen-zc}):
\begin{equation} \label{eq:cohen-zc}
    d_c = \frac{\beta_c}{\sqrt{\sigma_x^2 + \sigma_{\varepsilon}^2}},
\end{equation}
where $c = 1, \dots,3 $ and $\beta_c$ represents the fixed parameters estimates of $Z_c$, as in~(\ref{eq:mma}), and $\sigma_x$ and $\sigma_{\varepsilon}$ represent the variance of the random components and the error term, respectively, as expressed in~(\ref{eq:random-sigma}) and~(\ref{eq:error-sigma}).

%% file: paper/4_results.tex
\section{Results} \label{sec:results}

This section compares the performance results of the meta-learner proposed in this study with UOMS. Subsequently, a mixed model analysis of the factors contributing to the algorithm choice within the eight meta-learners $M_l$ is given. 

\subsection{Proposed approach versus UOMS}

\begin{table*}
\centering
\begin{threeparttable}
\caption{Comparison of the mean performance $\overline{y}^\texttt{SEL}$ and mean errors $\overline{D}$ with standard deviations across test datasets of two analysed meta-learner approaches, for AUC and AP.}
\label{tab:compare-two}
\begin{tabular}{llrrrrrr}
\toprule
 &  & Current study & UOMS & D.f. & T-stat. & p-value & Effect size \\ \midrule
\multirow{2}{*}{AUC} & $\overline{y}^\texttt{SEL}$ & \textbf{0.6703} $\pm$ 0.1820 & 0.6464 $\pm$ 0.1940 & 2317 & 7.882 & \textless{}0.001 & 0.126 \\
 & $\overline{D}$ & \textbf{0.1567} $\pm$ 0.1315 & 0.1804 $\pm$ 0.1558 & 2317 & -7.882 & \textless{}0.001 & 0.162 \\ \midrule
\multirow{2}{*}{AP} & $\overline{y}^\texttt{SEL}$ & \textbf{0.1413} $\pm$ 0.1955 & 0.1369 $\pm$ 0.1905 & 2302 & 2.193 & 0.028 & 0.009 \\
 & $\overline{D}$ & \textbf{0.1685} $\pm$ 0.1466 & 0.1749 $\pm$ 0.1543 & 2302 & -2.193 & 0.028 & 0.042 \\ \bottomrule
\end{tabular}
\begin{tablenotes}
\item D.f. -- degrees of freedom, T-stat. -- t-test statistic
\end{tablenotes}
\end{threeparttable}
\end{table*}

The performance comparison of the two solutions is presented in Table~\ref{tab:compare-two}. It lists the mean performance $\overline{y}^\texttt{SEL}$ measured as AUC and AP obtained over the set of test datasets $x_i$, and the meta-learners' mean error $\overline{D}$ as defined in~(\ref{eq:distance}). In both cases, AUC and AP, the solution proposed in the current study has a significantly higher mean performance than UOMS ($p<0.001$ for AUC, $p=0.028$ for AP).

\subsection{Statistical analysis}

\begin{table}
\centering
\begin{threeparttable}
\caption{Mean performance $\overline{y}^\texttt{SEL}$ with standard deviations across test datasets of eight meta-learners for AUC and AP.}
\begin{tabular}{p{0.13\linewidth}p{0.06\linewidth}p{0.08\linewidth}p{0.2\linewidth}p{0.2\linewidth}}
\toprule
Meta-model & AD alg. & Meta-features & $\overline{y}^\texttt{SEL}$, AUC & $\overline{y}^\texttt{SEL}$, AP \\ \midrule
$M_1$ (NN) & $A_2$ & $F_2$ & \textbf{0.692} $\pm$ 0.177 & \textbf{0.154} $\pm$ 0.215 \\
$M_1$ (NN) & $A_2$ & $F_1$ & 0.679 $\pm$ 0.179 & 0.150 $\pm$ 0.209 \\
$M_1$ (NN) & $A_1$ & $F_2$ & 0.678 $\pm$ 0.183 & 0.146 $\pm$ 0.200 \\
$M_1$ (NN) & $A_1$ & $F_1$ & 0.670 $\pm$ 0.182 & 0.141 $\pm$ 0.195 \\
$M_2$ (CF) & $A_2$ & $F_2$ & 0.646 $\pm$ 0.194 & 0.137 $\pm$ 0.191 \\
$M_2$ (CF) & $A_2$ & $F_1$ & 0.630 $\pm$ 0.202 & 0.136 $\pm$ 0.192 \\
$M_2$ (CF) & $A_1$ & $F_2$ & 0.630 $\pm$ 0.191 & 0.127 $\pm$ 0.176 \\
$M_2$ (CF) & $A_1$ & $F_1$ & 0.616 $\pm$ 0.193 & 0.120 $\pm$ 0.169 \\ \bottomrule
\end{tabular}%
\label{tab:mean-auc-ap}
\end{threeparttable}
\end{table}

\begin{table}
\centering
\begin{threeparttable}
\caption{Type III analysis of the main effects of the meta-learner components.}
\label{tab:effect-size}
\begin{tabular}{p{0.05\linewidth}p{0.23\linewidth}p{0.12\linewidth}p{0.12\linewidth}p{0.11\linewidth}p{0.07\linewidth}}
\toprule
 & Component & D.f. & F-stat. & p-value & Effect size \\ \midrule
\multirow{4}{*}{AUC} & Intercept & 1,2291 & 13382.78 & \textless{}0.001 & \\
 & Meta-features $F_q$ & 1,16244 & 51.457 & \textless{}0.001 & 0.082 \\
 & AD models $A_r$ & 1,16244 & 127.648 & \textless{}0.001 & 0.130 \\
 & Meta-model $m_p$ & 1,16244 & 678.938 & \textless{}0.001 & \textbf{0.300} \\ \midrule
\multirow{4}{*}{AP} & Intercept & 1,2294 & 14091.410 & \textless{}0.001 & \\
 & Meta-features $F_q$ & 1,16009 & 0.139 & 0.710 & 0.004 \\
 & AD models $A_r$ & 1,16025 & 69.087 & \textless{}0.001 & 0.095 \\
 & Meta-model $m_p$ & 1,16016 & 111.051 & \textless{}0.001 & 0.121 \\ \bottomrule
\end{tabular}
\begin{tablenotes}
\item D.f. -- degrees of freedom, F-stat. -- F-test statistic
\end{tablenotes}
\end{threeparttable}
\end{table}

A performance summary of eight meta-learners is presented in Table~\ref{tab:mean-auc-ap}. Interestingly, the hybrid approach of $F_2$, $A_2$, and $m_1$ gives the best performance results, for both meta-learner series, AUC and AP-based.

Table~\ref{tab:effect-size} presents the Type III main effects for the mixed model analysis.
Whereas the choice of each component $Z_c$ is statistically significant ($p<0.001$), the choice of the meta-model has the largest effect size. This is particularly visible when using AUC as a performance metric.
The AP metric did not demonstrate a notable difference due to its ``lower resolution'' (highly skewed distributions: skewness -- 2.647, kurtosis -- 7.064). The skewed distribution of AP values is expected as unbalanced datasets generally show this behaviour~\cite{ma2013imbalanced, viola2022metaap}. The NN meta-learner did, however, demonstrate a favourable performance in comparison to the CF group (Table~\ref{tab:mean-auc-ap}).

The interaction terms between the main effects outlined in~(\ref{eq:mma}) had no statistical significance and were subsequently excluded from the final statistical model.

\subsection{Time analysis}
Time analysis was performed on a subset of 20 datasets chosen at random from the entire set used in the current study. The datasets ranged in size from 68 to 5,186 observations and 5 to 147 features. This analysis has been restricted to the end-user perspective, which includes the generation of dataset meta-features and the prediction of the best-suited algorithm. The time summary of both approaches to generate meta-features and perform prediction is presented in Table~\ref{tab:time-mf} and Table~\ref{tab:time-predict}, respectively.

\begin{table}
\centering
\begin{threeparttable}
\caption{Time in seconds to generate dataset meta-features summarised for a random sample set of 20 datasets.}
\label{tab:time-mf}
\begin{tabular}{p{0.3\linewidth}p{0.1\linewidth}p{0.15\linewidth}}
\toprule
Statistic (time, s) & UOMS  & Current study \\ \midrule
Mean & 0.886 & 2.265 \\
Standard deviation & 0.777 & 3.627 \\
Min & 0.351 & 0.044 \\
Max & 3.103 & 10.517 \\ \midrule
Cases with shorter time & 8 & 12 \\ \bottomrule
\end{tabular}
\end{threeparttable}
\end{table}

\begin{table}
\centering
\begin{threeparttable}
\caption{Time in seconds to predict the best performing algorithm summarised for a random sample set of 20 datasets and across the meta-learner variants.}
\label{tab:time-predict}
\begin{tabular}{p{0.3\linewidth}p{0.1\linewidth}p{0.1\linewidth}}
\toprule
Statistic (time, s) & CF & NN \\ \midrule
Mean & 1.376 & 0.492 \\
Standard deviation & 0.574 & 0.114 \\
Min & 0.914 & 0.400 \\
Max & 4.537 & 0.956 \\ \bottomrule
\end{tabular}
\end{threeparttable}
\end{table}

Although the UOMS approach takes less time on average to generate the meta-feature set, the number of datasets for which the generation takes less time is greater for the currently presented approach. The current approach is more time-consuming for datasets with a relatively higher number of observations because it involves calculating the distances between all instances within a dataset. Ultimately, the time difference between UOMS and the current approach was not statistically significant for the measured sample set ($t_{19} = 1.835$, $p=0.082$).

When compared to the CF meta-model, the prediction times are shorter on average and more consistent with the use of NN. Furthermore, training times for CF models were significantly longer than for NN-based models (approx. 20 hours versus approx. 10 minutes, per meta-learner variant).

The above analysis was carried out on the machine with the following subcomponents: 1.6 GHz Dual-Core Intel Core i5 processor and 8 GB of 2133 MHz RAM.

%% file: paper/5_discussion.tex
\section{Discussion} \label{sec:discussion}
This research performed a direct comparison between a new method proposed here with the UOMS approach~\cite{metaod2021}. The results demonstrate that while there was a statistically significant improvement with the proposed method, there was a negligible effect (practical difference) when comparing the AP mean error (Cohen's $d$ = 0.042) and a small effect when comparing the AUC mean error (Cohen's $d$ = 0.162) between the two approaches (Table~\ref{tab:compare-two}). The proposed method in this study demonstrates, however, that equivalent results can be obtained from a substantially reduced feature set and the omission of HO in the meta-learning configurations. Previous works have assumed that HO was the main characteristic in meta-learning. Whereas the initial results are only a direct comparison between the method proposed in the current study and the UOMS, the mixed model analysis helps to elucidate the characteristics that have a contributing effect. 

The mixed model analysis, summarised in Table~\ref{tab:effect-size}, shows that for AUC, the larger meta-feature set ($F_2$) provides only a marginal benefit over the smaller set of meta-features $F_1$ (Cohen's $d$ = 0.082). It is worth noting that while $F_2$ makes extensive use of generic statistical features, the compact set $F_1$ is crafted to reflect anomaly characteristics.
When comparing differences in the performance between the two sets of AD models, the large set with HO ($A_2$) outperforms the small set without HO ($A_1$). However, given the number of models in both groups (298 versus 13), the effect size is not as compelling as one would expect (Cohen's $d$ = 0.130).
The analysis demonstrates that the choice of meta-model has the most significant impact on the meta-learner's final performance, with the NN-based meta-model $m_1$ outperforming the state of the art CF approach $m_2$ (Cohen's $d$ = 0.300).

This outcome is an important consideration given that current AutoML or meta-learning studies frequently direct their attention to other aspects, such as meta-features development~\cite{metafeatures_general2016,ad_metafeatures2021} or HO~\cite{komer2014hyperopt,bayesian_hyperparam_opt2015,hyperparam_opt2016}.
This work demonstrates that investing time and effort into creating an adequate meta-model that can successfully utilise data from historical evaluations is the most promising approach for improving meta-learners for unsupervised AD.

The contributions measured on AP show a similar pattern, however, the effects are less visible. With a significance level at 0.05, the influence of the meta-features is not statistically significant. Consequently, the effect size is negligible. The contributions of the other two components are larger, but their effect sizes on the AP metric are also minimal. 

The results of the time analysis show that using a meta-learner within an AD pipeline outweighs the costs in terms of time and computing resources. For a dataset with 1,000 observations and 45 features, the extra time of 1-2 seconds for meta-feature generation and 0.5 seconds for finding the best suited algorithm could potentially save hours on a trial-and-error process of finding the best performing algorithm and evaluating the results.

%% file: paper/6_conclusions.tex
\section{Conclusion} \label{sec:conclusions}
In this study, a new meta-learner to help in the identification of an appropriate unsupervised machine learning algorithm when applied to anomaly detection problems was presented. When compared to the current state of the art, UOMS~\cite{metaod2021}, the proposed method demonstrated a statistically significant improvement in results using two performance metrics, AUC and AP. 

In addition, using a $2^3$ experimental design, an experiment was conducted to understand which of the component parts (meta-model, meta-features, algorithm set) in both approaches had the greatest influence on the overall performance of the meta-learner. While the choice of meta-features and the base set of AD algorithms were shown to have a relatively small effect size, the meta-model choice had the largest effect on the meta-learner performance. Furthermore, the analysis revealed that a hybrid version of both approaches, UOMS and the one proposed here, gave the best performance results. 

Finally, the experiments in this study were carried out on the largest number of datasets used to date for examining an algorithm selection for an unsupervised anomaly detection task. 

Future work in this area should focus on an in-depth analysis of approaches used in the meta-models and the development of datasets that represents a truly robust test to any future meta-learners.